\documentclass[10pt,twocolumn,letterpaper]{article}
\usepackage[pagenumbers]{cvpr}
\usepackage{times,graphicx,amsmath,amssymb,booktabs}
\usepackage{float}
\definecolor{cvprblue}{rgb}{0.21,0.49,0.74}
\usepackage[pagebackref,breaklinks,colorlinks,allcolors=cvprblue]{hyperref}
\definecolor{citeorange}{HTML}{D9730D}
\hypersetup{citecolor=citeorange} 
\usepackage{colortbl} 
\usepackage{amsmath,booktabs,array,tabularx}
\definecolor{RLGain}{rgb}{0.89,0.96,0.90}
\definecolor{RLDrop}{rgb}{0.99,0.91,0.91}
\newcommand{\rlgain}[1]{\cellcolor{RLGain}$#1$}
\newcommand{\rldrop}[1]{\cellcolor{RLDrop}$#1$}

\newcommand{\sg}[1]{\operatorname{sg}\!\left[#1\right]}

\definecolor{wacW}{HTML}{1F5FA8}   
\definecolor{wacA}{HTML}{13866F}   
\definecolor{wacC}{HTML}{7B3FA0}   
\definecolor{OursBG}{HTML}{E8F1FA} 

\newcommand{\ours}{\rowcolor{OursBG}\cellcolor{white}} 

\newcommand{\W}{\textcolor{wacW}{W}}
\newcommand{\A}{\textcolor{wacA}{A}}
\newcommand{\C}{\textcolor{wacC}{C}}
\newcommand{\dexwac}{Dex\W\A\C}
\usepackage{multirow}

\title{\dexwac: A Unified {\W}orld--{\A}ctor--{\C}ritic Model\\
for Dexterous Robot Manipulation}
\author{%
Hongjia Liu$^{1,2}$\quad Dacheng Zhou$^{2,3}$\quad Minghao Fu$^{1}$\quad Chuanjun Guo$^{2}$\\
Tianze Wang$^{2}$\quad Qingtao Liu$^{2}$\quad He Wang$^{2,4}$\quad Li Yi$^{2,5,6}$\\[3pt]
{\normalsize $^{1}$University of California, San Diego\quad  $^{2}$Galbot\quad $^{3}$École Polytechnique Fédérale de Lausanne (EPFL), Switzerland}\\
{\normalsize $^{4}$Peking University\quad $^{5}$Tsinghua University\quad $^{6}$Shanghai Qi Zhi Institute}%
}
\begin{document}
\maketitle
\begin{abstract}
Robot demonstrations provide more than actions to imitate: they reveal how actions change the world and how those changes contribute to task success. Recent world--action models (WAMs) extend vision--language--action (VLA) models by jointly predicting actions and future observations, yet this predictive ability serves only as an auxiliary loss for policy learning. Meanwhile, most policy improvement methods learn dynamics and values in separate models, increasing training overhead and limiting representation sharing between these complementary objectives. We introduce \textbf{\dexwac}, a unified world--actor--critic model that jointly learns an actor, an action-conditioned world model, and a bootstrapped critic in a single robot pretraining stage. Our key insight is that visual prediction encourages control-relevant features, while value supervision promotes task-relevant representations; together, they shape a representation suited to policy learning.
Built on an image-editing backbone, DexWAC predicts, for each action chunk, the visual latent and proprioception at its endpoint together with its reward and continuation, so the world model alone efficiently yields complete imagined transitions without generating intermediate video frames.
After adapting the world model and critic on real rollouts, we improve the pretrained actor through either critic-guided best-of-$N$ selection and PPO on imagined rollouts. DexWAC surpasses the large-scale pretrained VLA and WAM baselines in simulation and
achieves 75\% average success on two real-world tasks, compared with 45\% for $\pi_{0.5}$. Its world model is more accurate and faster than video-based baselines, and ablations confirm that learning dynamics and value together improves the policy. \small{\href{https://lhj-lhj.github.io/DexWAC/}{Project~page.}}
\end{abstract}

\begin{figure*}[t]
\centering
\includegraphics[width=\textwidth]{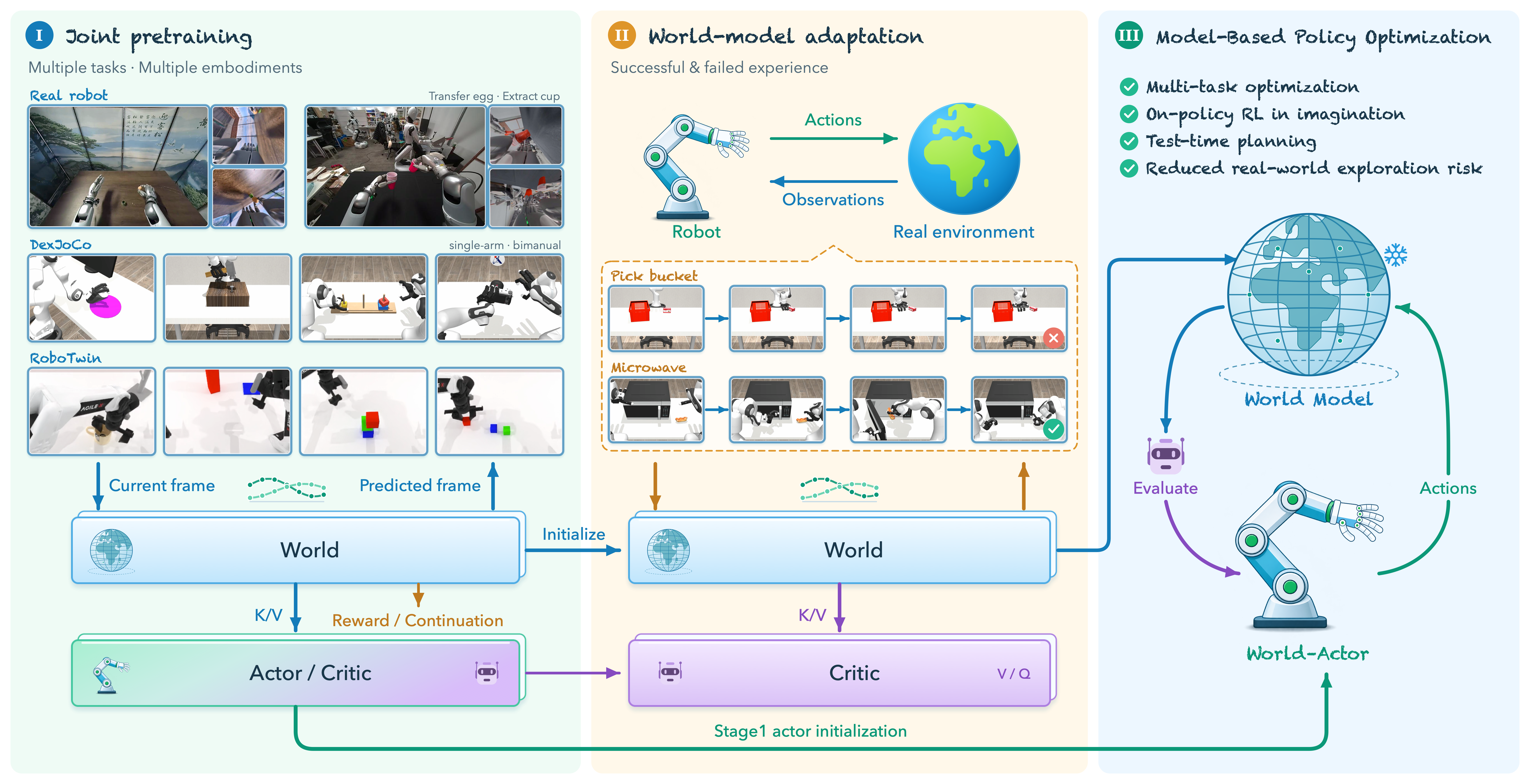}
\caption{\textbf{Overview of DexWAC.}
\textbf{(I)} Unified world--actor--critic pretraining on
existing robotic datasets, optionally incorporating
non-expert trajectories for dynamic and value learning.
\textbf{(II)} After the pretrained actor executes,
the world model and critic are adapted to the resulting
successful and failed trajectories, while the actor
remains unchanged. \textbf{(III)} With the adapted world-model backbone frozen,
the actor initialized from Stage~I and the critic adapted
in Stage~II support policy improvement through imagined
interactions across tasks and embodiments.
Snowflakes and flames indicate frozen and trainable
components, respectively.}
\label{fig:workflow}
\end{figure*}

\section{Introduction}
Large-scale learning from demonstrations has enabled robot
policies to operate across tasks and embodiments,
offering a scalable path toward general-purpose physical
intelligence~\cite{black2024pi_0,intelligence2025pi_,
nvidia2026cosmos3}. Yet imitating actions alone leaves much of the supervision in these demonstrations unused. 
A demonstration reveals not only which action an expert
chose, but also how it changes the
world~\cite{wu2024unleashing} and how those changes
contribute to task success~\cite{zhang2023interpretable}.
Data collection and policy execution also yield suboptimal trajectories, failed attempts, and unstructured play, which are often underused by imitation-based policies.
Exploiting this supervision can help a policy understand
both \emph{what its actions can change} and
\emph{which factors matter for the task}~\cite{liu2023learning}.

Recent world--action models (WAMs) jointly predict actions and future observations~\cite{bi2025motus,guo2026video}, but use this prediction mainly as an auxiliary objective for policy learning.
Meanwhile, methods to improve pretrained policies often introduce separately trained world models~\cite{guo2026ctrlworld, zhang2026vlambpo} and reward or value models~\cite{lee2026roboreward,
wang2026world}.
For a generalist policy, these models must themselves generalize across
the same broad range of tasks and embodiments as the policy~\cite{liang2026robometer}.
Pretraining them separately at this scale can duplicate computation and relearn overlapping representations.
It also keeps the predictive and value objectives from jointly shaping the features the actor uses~\cite{cobbe2021ppg}.
These observations motivate two questions: \emph{Can a single pretraining stage jointly acquire an actor, a world model, and a critic?
And can their complementary objectives constrain a shared representation that improves policy learning?}

We therefore introduce \textbf{\dexwac} (Fig.~\ref{fig:workflow}), a unified {\W}orld--{\A}ctor--{\C}ritic model for dexterous manipulation that learns all three components in a single pretraining stage.
To share one representation across these components, we adopt a Mixture-of-Transformers (MoT) architecture~\cite{liang2025mot} that couples a generative world model backbone with a compact actor--critic expert (Fig.~\ref{fig:architecture}).
The backbone encodes the language instruction, current visual
observations, and proprioception as a shared prefix of attention keys
and values (K/V). During joint training, action-conditioned forward
prediction encourages this prefix to capture how an action chunk
changes the scene and robot state, while value supervision uses
bootstrapped temporal-difference targets to shape the same prefix
around progress toward task success. Through action imitation, the
actor learns to use these complementary control- and task-relevant
features to generate expert action chunks (Fig.~\ref{fig:attention}).

To enable efficient imagination, we build \textbf{\dexwac}
on an image-editing backbone with rich visual priors. It predicts the visual latent
and proprioceptive state at the end of each action chunk without generating intermediate frames. These predictions provide the context
for the next imagined chunk. After pretraining, we further adapt the world model and critic on successful and failed rollouts collected by the pretrained actor. We then use the adapted world model for model-based policy
improvement in two ways~\cite{janner2019mbpo}. 
For critic-guided best-of-$N$ (BoN), we sample $N$
candidate action chunks from the frozen actor at inference time. The adapted critic then ranks them using their
world-model-predicted endpoints and executes the highest-valued chunk. For PPO~\cite{schulman2017ppo}, we use the world model to generate imagined rollouts and update the actor directly on them.

We evaluate DexWAC on two simulation benchmarks, DexJoCo~\cite{wang2026dexjoco} and RoboTwin~2.0~\cite{chen2025robotwin2}, and on two contact-rich manipulation tasks using a physical dexterous robot.
Joint pretraining alone yields a policy that reaches 65.5\%
success on DexJoCo, compared with 57.9\% for $\pi_{0.5}$~\cite{intelligence2025pi_},
and the best average success on RoboTwin~2.0 (88.1\%).
Its world model achieves the best prediction quality on all
six image-quality metrics, reducing wrist-view LPIPS by 38\% while running $8.5\times$ faster than the strongest baseline. Using only 50 executions per task for adaptation, PPO in
imagination raises DexJoCo success to 67.8\%, surpassing the
large-scale pretrained WAMs, and improves bimanual success from 47.8\% to 52.9\%.
On a physical robot, DexWAC outperforms $\pi_{0.5}$ on two contact-rich tasks (50\% vs.\ 20\% and 100\% vs.\ 70\%).
Ablations further show that removing either world prediction or value supervision reduces policy success. \textbf{In summary, our contributions are as follows:}

\begin{enumerate}
    \item We propose \dexwac, which jointly trains an actor,
    a world model, and a critic with shared representations
    in a single pretraining stage.

    \item We evaluate the pretrained policy in two simulations
    and on two contact-rich dexterous manipulation tasks
    with a physical robot, and demonstrate the benefits
    of joint supervision through ablations.

    \item We adapt the world model and critic to execution
    data, evaluate world-model accuracy and efficiency,
    and demonstrate their utility for multi-task policy
    improvement through imagined PPO and critic-guided BoN.
\end{enumerate}
\section{Related Work}

\subsection{Unified Foundation Models}
Unified foundation models learn shared representations across multiple
objectives, enabling mutually beneficial transfer.
Transfusion~\cite{zhou2025transfusion},
Show-o~\cite{xie2025showo}, and BAGEL~\cite{deng2025bagel}
demonstrate the benefits of jointly learning multimodal understanding
and generation.
Embodied models extend this paradigm to action generation,
often leveraging visual generative
priors~\cite{guo2026video,zhang2026imagewam}.
UWM~\cite{zhu2025uwm}, Cosmos~3~\cite{nvidia2026cosmos3},
and Motus~\cite{bi2025motus} unify understanding, future prediction,
and action generation, but primarily use dynamics prediction as
auxiliary supervision.
LeWAM~\cite{fu2026lewam} uses the learned dynamics at
test time to evaluate sampled actions, but learns neither rewards nor values.
FACT~\cite{peng2026fact} and Cosmos
Policy~\cite{kim2026cosmospolicy} further incorporate value prediction
into world--action modeling.
However, they supervise value prediction with trajectory-derived
normalized progress labels, rather than learning a
temporal-difference critic for policy improvement.
Shared policy–value representations have also been explored in
reinforcement learning (PPG~\cite{cobbe2021ppg}) and language agents
\cite{liang2026sapo}.
DexWAC extends this unification to joint learning of action generation, action-conditioned world modeling, reward prediction, and
temporal-difference value estimation through shared representations.

\subsection{Policy Improvement for Pretrained Policies}

Pretrained policies can fail on states poorly covered
by their demonstrations, motivating further learning
from suboptimal and failed executions.
Online RL addresses this gap through
environment interaction~\cite{chen2025pirl,wagenmaker2025dsrl},
but real-world exploration is slow and costly,
especially across multiple tasks and embodiments.
Offline RL learns from existing data, but unreliable value estimates
can destabilize policy optimization and lead to policy
collapse~\cite{kumar2020cql,kostrikov2022iql}.
V-GPS~\cite{BoN} and SeeQ~\cite{singh2026seeq} instead learn critics
from diverse robot data and use them to rank candidate actions at test
time, steering behavior without directly updating the policy.
World models offer an alternative by learning environment dynamics
and generating imagined experience for policy optimization and
value learning~\cite{zhang2026vlambpo,park2026mac,guo2026vlaw, hafner2025dreamerv3}.
However, achieving broad dynamics coverage often
requires separate world-model pretraining and
adaptation, introducing additional training
costs~\cite{guo2026ctrlworld}.
DexWAC instead adapts the dynamics, reward model, and critic
learned during joint pretraining and reuses them for policy
optimization, reducing the need
for a separately trained world model.
\begin{figure*}[t]
    \centering
    \includegraphics[width=\textwidth]{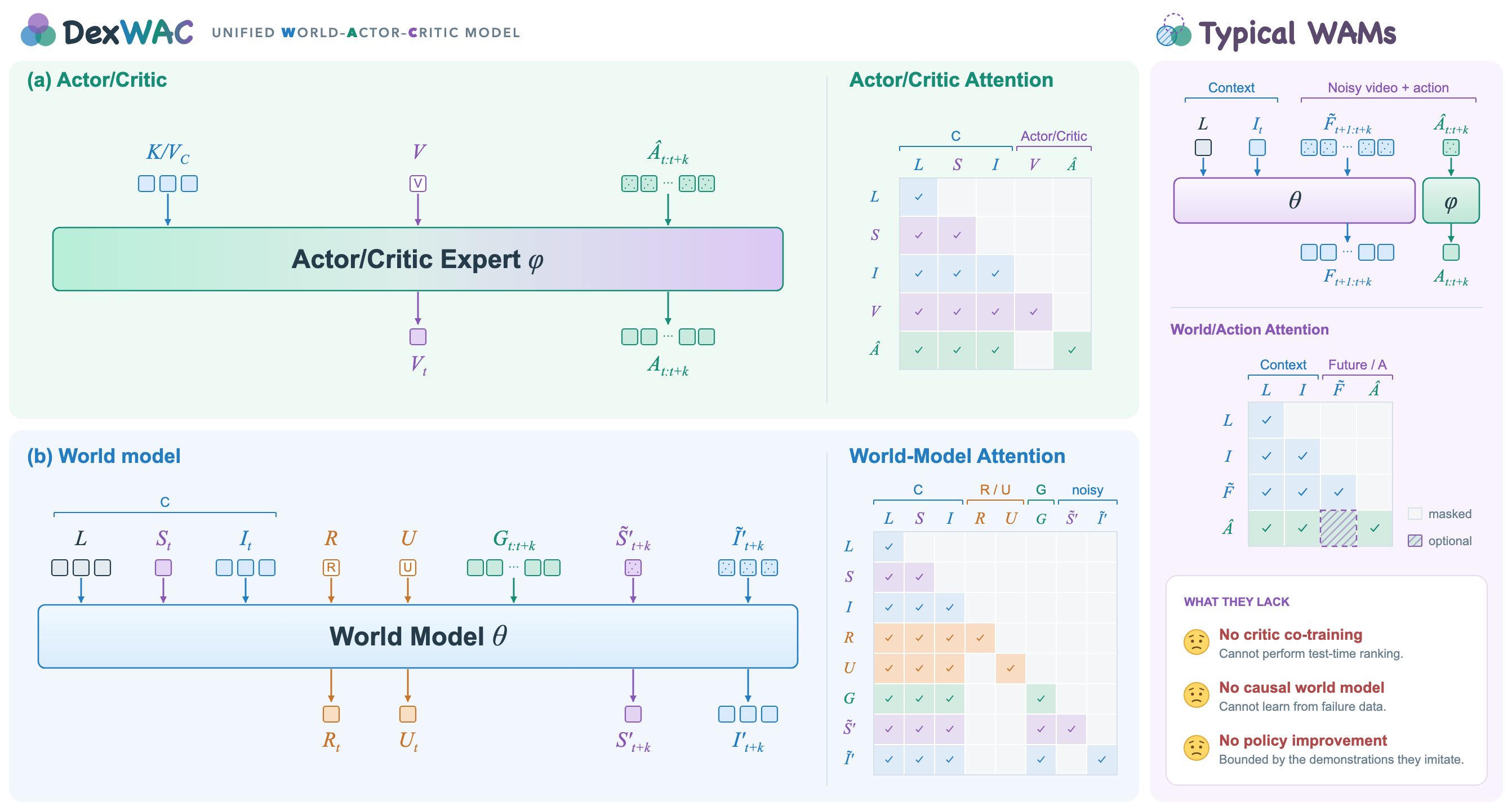}

\caption{\textbf{DexWAC architecture.}
\textbf{(a) Actor/critic.}
The actor--critic expert $\phi$ reads contextual keys
and values from the world backbone to generate action
chunks and estimate state values. Attention masks keep
value prediction action-independent; optional clean-action
tokens $G$ and a $Q$ token enable action-value prediction
for test-time ranking.
\textbf{(b) World model.}
The backbone $\theta$ predicts visual latents and
proprioception at the action-chunk endpoint, conditioned
on the observed context and clean actions. Reward and
continuation depend only on the observed context.
\textbf{Right: WAM comparison.}
Typical WAMs jointly predict video and actions, without
explicit critic co-training or action-conditioned dynamics
for policy improvement. Attention rows and columns denote
queries and keys/values, respectively.}
    \label{fig:architecture}
\end{figure*}

\section{Preliminaries}
\label{sec:preliminaries}

\paragraph{Language-conditioned robot control.}
We formulate robotic manipulation as a language-conditioned
MDP $\mathcal{M}=(\mathcal{X},\mathcal{A},\mathcal{L},
P,r,\rho_0,\gamma)$, where $\mathcal{X}$, $\mathcal{A}$,
and $\mathcal{L}$ are the state, action, and instruction
spaces, respectively;
$P(x_{t+1}\mid x_t,a_t)$ denotes the transition dynamics,
$r$ the reward function, $\rho_0$ the initial-state
distribution, and $\gamma$ the discount factor.
Given state $x_t$ and instruction $\ell$, the policy
samples an action chunk
$\mathbf a_t=(a_t,\ldots,a_{t+H-1})\sim\pi(\cdot\mid x_t,\ell)$.
The objective is to maximize the expected discounted return:
\begin{equation}
    J(\pi)=
    \mathbb{E}_{\ell\sim p(\ell),\,\xi\sim(\rho_0,\pi,P)}
    \left[\sum_{t=0}^{T-1}\gamma^t r_t\right],
    \label{eq:control-objective}
\end{equation}
where $p(\ell)$ is the instruction distribution,
$T$ is the episode length, and
$r_t=r(x_t,a_t,x_{t+1},\ell)$ is the reward at step $t$.

\section{Method}
\label{sec:method}

DexWAC proceeds in three stages (Fig.~\ref{fig:workflow}).
Joint pretraining learns, from the same demonstrations, an actor for action generation,
a world model for predicting the visual and proprioceptive
consequences of actions, and a critic for state-value
estimation (Sec.~\ref{sec:pretrain}). Next, we adapt the world model and critic on real rollouts of the pretrained actor, so that they remain
accurate on the states the policy actually visits (Sec.~\ref{sec:adapt}). Finally, the adapted components improve control, either by updating the actor in imagination or by ranking its proposals (Sec.~\ref{sec:imagine}). Hyperparameters and implementation details are
given in Appendix~\ref{app:hyperparameters}.

\subsection{Learning to Act, Predict, and Evaluate}
\label{sec:pretrain}



We design the pretraining stage so that one representation serves
acting, predicting, and evaluating. A Mixture-of-Transformers (MoT)
architecture~\cite{liang2025mot} couples a generative world backbone
$\theta$ with a compact actor--critic expert $\phi$
(Fig.~\ref{fig:architecture}). The backbone encodes the context
$C_t=(\ell,z_t,s_t)$ of instruction, visual latents
$z_t=\mathcal{E}_{\mathrm{VAE}}(o_t)$~\cite{rombach2022high}, and
proprioceptive state, and the expert reads its keys and values (K/V).
The two thus share features through attention but keep separate
weights, which limits interference with the backbone's pretrained
visual priors~\cite{deng2025bagel}.

Two choices make the backbone a complete simulator for imagination.
First, given $C_t$ and an action chunk, it predicts only the visual
latent and proprioceptive state at the chunk endpoint
$t+H$~\cite{park2026mac}, so each imagined step needs a single
prediction. Second, it predicts reward and continuation from the reached state, so each imagined transition, including its feedback, comes entirely from the world model.

\paragraph{Action and future-state prediction.}
The actor learns expert action chunks $\mathbf a_t$
through behavior cloning~\cite{pomerleau1989alvinn}, while the world backbone
predicts $(z_{t+H},s_{t+H})$ from $C_t$ and clean actions,
with information flow controlled by the attention masks
in Fig.~\ref{fig:architecture}. All three continuous targets are trained with conditional flow matching~\cite{lipman2023flow}: given a clean target $x_0$ and Gaussian noise $\epsilon$, we form $x_\tau=(1-\tau)x_0+\tau\epsilon$ and regress the velocity
\begin{equation}
    \mathcal{L}_{\mathrm{FM}}
    = \mathbb{E}_{\tau,\epsilon}
      \big\|v_\theta(x_\tau,\tau)-(\epsilon-x_0)\big\|_2^2 .
    \label{eq:action-flow}
\end{equation}
Applied to actions, visual latents, and proprioceptive states, this yields
$\mathcal{L}_A$, $\mathcal{L}_I$, and $\mathcal{L}_S$, respectively.

\paragraph{Reward prediction.}
Robotic tasks often provide sparse success feedback,
while dense rewards are costly to specify and
annotate~\cite{rocamonde2024vlmrewards}.
To provide denser supervision for reward learning,
we draw on DreamSmooth~\cite{lee2024dreamsmooth}
and distribute terminal success feedback over the
preceding $D_T=\min(W_R,T)$ steps using a one-sided
Gaussian kernel:
\begin{equation}
\begin{aligned}
    \widetilde r_j
    &= \frac{w_{T-j}}{Z_T}
       \mathbb{I}[0\leq T-j\leq D_T], \\
    w_k &= \exp\!\left(-\frac{k^2}{2\sigma^2}\right),
    \qquad Z_T=\sum_{k=0}^{D_T}w_k.
\end{aligned}
\end{equation}
Here $k$ denotes the temporal distance from the successful
terminal observation, and $\sigma$ controls the Gaussian
bandwidth. Normalization yields unit total reward per
successful episode; failed episodes retain zero rewards.
For a chunk transition from $t$ to $e=\min(t+H,T)$,
let $d=e-t$ denote the number of executed control steps.
Its target is the discounted reward accumulated on arrival:
\begin{equation}
    R_{t:e} = \textstyle\sum_{i=1}^{d} \gamma_{\text{step}}^{\,i-1}\,
    \widetilde r_{t+i}.
\label{eq:intervalreward}
\end{equation}
The reward head $\widehat R(C_e,d)$ sees only the endpoint context and the
interval length, so it can be applied directly to predicted endpoints during
imagination. Since positive targets remain sparse, we balance reward
regression by averaging squared errors separately over
positive and zero targets:
\begin{equation}
    \mathcal{L}_R
    = \frac{\alpha}{|\mathcal{P}|}\sum_{i\in\mathcal{P}}
      (\widehat R_i - R_i)^2
    + \frac{1-\alpha}{|\mathcal{N}|}\sum_{i\in\mathcal{N}}
      (\widehat R_i - R_i)^2 ,
\label{eq:rewardbalance}
\end{equation}
with $\mathcal{P}=\{i:R_i>0\}$ and $\mathcal{N}=\{i:R_i=0\}$ in each update,
falling back to plain MSE if either set is empty.

\paragraph{Continuation prediction.}
The continuation head predicts
$\widehat c_t=\operatorname{sigmoid}(u_t)$ from $C_t$,
where the binary label $c_t$ indicates whether value
bootstrapping continues. We also balance the two classes as in reward regression:
\begin{equation}
\begin{aligned}
    \mathcal{L}_U
    ={}&-\frac{1}{2|\mathcal{U}_1|}
         \sum_{i\in\mathcal{U}_1}\log\widehat c_i 
       &-\frac{1}{2|\mathcal{U}_0|}
         \sum_{i\in\mathcal{U}_0}\log(1-\widehat c_i).
\end{aligned}
\label{eq:continuation}
\end{equation}
Here $\mathcal{U}_b=\{i:c_i=b\}$ contains valid examples
across the complete optimizer update. If either class
is absent, we use ordinary binary cross-entropy over
the remaining class.

\paragraph{Value learning.}
With rewards and continuations defined, we learn the state value by chunk-level
temporal-difference (TD) learning~\cite{sutton1988td,park2025horizon} from the
start of pretraining, so that value estimation shapes the representation
rather than being introduced from scratch during policy optimization. Training
samples are organized as \emph{pairs}: each anchor context $C_t$ is batched
with the context $C_e$ at its observed endpoint $e=\min(t+H,T)$, and both pass
through the same network in a single forward pass. With $d=e-t$,
\begin{align}
    y_t &= R_{t:e} + \gamma_{\text{step}}^{\,d}\, c_e\,
           \operatorname{sg}\!\left[V_\phi(C_e)\right],
    \label{eq:valuetarget}\\
    \mathcal{L}_V &= \mathbb{E}_{t}\!\left[
        h_\kappa\!\left(V_\phi(C_t)-y_t\right)\right],
    \label{eq:td}
\end{align}
where $\operatorname{sg}$ denotes stop-gradient and $h_\kappa$ the Huber loss~\cite{huber1964robust}. The bootstrap value is obtained from the current network at the paired endpoint and detached, eliminating the need for a separate target network or EMA copy. Critic learning thus adds minimal computational overhead, requiring only an additional value token and prediction head.


\paragraph{Joint objective.}
The pretraining objective combines all six terms:
\begin{align}
    \mathcal{L}_{\mathrm{sup}}
    ={}& \lambda_A\mathcal{L}_A
       + \lambda_I\mathcal{L}_I
       + \lambda_S\mathcal{L}_S \nonumber\\
       &+ \lambda_R\mathcal{L}_R
       + \lambda_U\mathcal{L}_U
       + \lambda_V\mathcal{L}_V.
    \label{eq:sup}
\end{align}

\subsection{Model Adaptation with Real Rollouts}
\label{sec:adapt}

To cover states encountered during policy execution,
we adapt a copy of the pretrained world model and critic
on the expert demonstrations together with successful and
failed rollouts of the pretrained actor, keeping the
original actor unchanged.
Contextual keys and values remain attached, so predictive
and value losses jointly update the backbone and critic.
The adaptation objective is:
\begin{equation}
\begin{aligned}
    \mathcal{L}_{\mathrm{adapt}}
    ={}& \lambda_I\mathcal{L}_I
       + \lambda_S\mathcal{L}_S
       + \lambda_R\mathcal{L}_R \\
       &+ \lambda_U\mathcal{L}_U
       + \lambda_V\mathcal{L}_V.
\end{aligned}
\label{eq:adapt}
\end{equation}


\subsection{Policy Improvement through Imagination}
\label{sec:imagine}

The adapted model serves as a learned simulator in which the policy can be
improved without further environment
interaction~\cite{hafner2025dreamerv3,zhang2026vlambpo}. Rollouts are kept
short to limit compounding model error~\cite{janner2019mbpo} and operate
directly on visual latents and proprioceptive states, without decoding to RGB.

\paragraph{Imagined trajectories.}
Each rollout starts from a real observation (a \emph{root}), with roots
balanced across tasks and between expert and policy-rollout data
(Appendix~\ref{app:hyperparameters}).
At each step, the collection policy generates an action chunk
and the frozen world model predicts the next latent observation and robot
state $\widehat C_{j+1}$. Reward and continuation are read from this clean
predicted endpoint:
\begin{equation}
    \widehat R_j=\widehat R(\widehat C_{j+1},H),\quad
    \widehat c_j=\mathbb{I}
    \big[\sigma(\widehat u(\widehat C_{j+1}))\geq\tau_c\big].
    \label{eq:imagined-feedback}
\end{equation}
where $\sigma$ denotes the sigmoid function, $\tau_c$ is
the continuation threshold, and the reward is the raw
linear output. Each root yields up to $K$ transitions, and terminated transitions are masked. We use these rollouts in two ways.

\paragraph{Critic-guided best-of-$N$.}
\label{par:critic-guided-bon}
Following BoN~\cite{BoN}, we sample $N$ candidate action chunks
$\mathbf a^{(n)} \sim \pi_{\mathrm{pre}}(\cdot\mid C_t)$
from the frozen actor at the same observation $C_t$.
On RoboTwin~2.0, we rank these candidates directly with the
adapted world model and state-value critic. The world model
predicts the visual and proprioceptive state at the end of
each chunk, using the same noise $\xi_t$ for all candidates.
We execute the chunk with the highest predicted endpoint value
and replan from the next real observation:
\begin{equation}
\begin{aligned}
    \widehat C_{t+H}^{(n)}
    &= \mathcal W_\theta(C_t,\mathbf a^{(n)};\xi_t), \\
    n^\star
    &= \arg\max_{1\leq n\leq N}
       V_\omega(\widehat C_{t+H}^{(n)}).
\end{aligned}
\label{eq:bon}
\end{equation}
This procedure requires no further training after adaptation.
We use $N=4$ chunks of $H=16$ actions on RoboTwin tasks
where the pretrained actor achieves less than 85\% success
(Table~\ref{tab:robotwin}).

On DexJoCo, we further reduce test-time planning cost by distilling
the world model's predictions and the critic's value estimates
into an action-value function, following MAC~\cite{park2026mac}.
We train a $Q$ expert on imagined transitions with the
one-chunk target
\begin{equation}
    y_Q = \sg{\widehat R(\widehat C',H)
          + \Gamma\,\widehat c'\,V_\omega(\widehat C')},
    \label{eq:q-target}
\end{equation}
where $\widehat C'$ is the predicted endpoint, $\widehat c'$
is its continuation indicator, and $\Gamma$ is the chunk discount.
At inference, we rank 8 proposals by $Q_\psi(C_t,\mathbf a^{(n)})$
and execute the highest-scoring chunk. The $Q$ expert thus
scores each candidate without generating its future observation.





\paragraph{PPO for flow-based policies.}

We further test whether the actor can be directly optimized in the learned world model using the pretrained critic. As shown in Fig.~\ref{fig:workflow}, the actor retains its pretrained initialization, while the critic inherits the adapted model. 
With both backbones frozen, we update the actor expert
using PPO with BC regularization, and the critic expert
using value regression.
Chunk-level advantages $\widehat A$ are estimated with GAE~\cite{schulman2016gae}. 
Following $\pi_{\mathrm{RL}}$~\cite{chen2025pirl}, we compute
the PPO likelihood ratio $\rho$ for a selected stochastic
denoising transition and optimize
\begin{equation}
    \mathcal L_\pi
    =-\mathbb E\!\left[
      \min\!\left(\rho\widehat A,\,
      \operatorname{clip}_\epsilon(\rho)\widehat A\right)
    \right]
    +\beta\mathcal L_{\mathrm{reg}},
    \label{eq:ppo-anchor}
\end{equation}
where $\operatorname{clip}_\epsilon(\rho)
=\operatorname{clip}(\rho,1-\epsilon,1+\epsilon)$.
The BC regularization term $\mathcal L_{\mathrm{reg}}$
uses expert actions through Eq.~\eqref{eq:action-flow}
to anchor the policy to the pretraining data distribution.
Imagined states and learning targets are detached,
so actor optimization does not backpropagate through
the world model.
\begin{table*}[t]
\centering
\caption{
Success rates (\%) on the eight-task DexJoCo subset,
evaluated with seeds 0, 1, and 2 using 50 episodes per
task and seed. Entries report the mean
with the sample standard deviation across seeds. Individual seed results are reported in Table~\ref{tab:dexwac-per-seed} in the appendix. \textbf{Bold} and \underline{underline} indicate the highest and second-highest mean in each column, including ties;
DexWAC rows are shaded in blue.
}
\label{tab:main-comparison}
\setlength{\tabcolsep}{3pt}
\renewcommand{\arraystretch}{1.15}
\resizebox{\textwidth}{!}{%
\begin{tabular}{@{}l*{11}{c}@{}}
\toprule
& \multicolumn{5}{c}{Single-arm tasks}
& \multicolumn{3}{c}{Bimanual tasks}
& \multicolumn{3}{c}{Average} \\
\cmidrule(lr){2-6}
\cmidrule(lr){7-9}
\cmidrule(lr){10-12}
Method
& \shortstack{Click\\mouse}
& \shortstack{Fold\\glasses}
& \shortstack{Hammer\\nail}
& \shortstack{Pick\\bucket}
& \shortstack{Water\\plant}
& Hanoi
& \shortstack{Microwave\\cook}
& Photograph
& Single-arm
& Bimanual
& Overall \\
\midrule

$\pi_{0.5}$~\cite{intelligence2025pi_}
& $68.67_{\pm 4.2}$
& $52.00_{\pm 3.5}$
& $60.67_{\pm 9.0}$
& $79.33_{\pm 3.1}$
& $68.00_{\pm 12.0}$
& $\underline{20.67}_{\pm 7.0}$
& $73.33_{\pm 10.3}$
& $40.67_{\pm 8.1}$
& $65.73_{\pm 5.8}$
& $44.89_{\pm 8.0}$
& $57.92_{\pm 5.8}$ \\

Motus~\cite{bi2025motus}
& $\mathbf{77.33}_{\pm 9.5}$
& $59.33_{\pm 3.1}$
& $\mathbf{79.33}_{\pm 6.1}$
& $74.00_{\pm 3.5}$
& $\mathbf{94.67}_{\pm 3.1}$
& $17.33_{\pm 9.2}$
& $80.67_{\pm 8.1}$
& $\mathbf{52.00}_{\pm 10.6}$
& $\mathbf{76.93}_{\pm 2.7}$
& $50.00_{\pm 4.1}$
& $\underline{66.83}_{\pm 3.2}$ \\

FACT (Stage1)~\cite{peng2026fact}
& $61.33_{\pm 5.0}$
& $60.67_{\pm 3.1}$
& $16.00_{\pm 4.0}$
& $84.67_{\pm 5.8}$
& $78.00_{\pm 4.0}$
& $10.00_{\pm 4.0}$
& $64.00_{\pm 2.0}$
& $18.67_{\pm 9.0}$
& $60.13_{\pm 0.9}$
& $30.89_{\pm 4.4}$
& $49.17_{\pm 2.2}$ \\

FACT~\cite{peng2026fact}
& $70.00_{\pm 12.5}$
& $58.67_{\pm 3.1}$
& $13.33_{\pm 1.2}$
& $80.67_{\pm 5.0}$
& $90.00_{\pm 0.0}$
& $3.33_{\pm 1.2}$
& $62.67_{\pm 1.2}$
& $30.00_{\pm 6.0}$
& $62.53_{\pm 3.6}$
& $32.00_{\pm 2.4}$
& $51.08_{\pm 3.1}$ \\

\midrule

\ours DexWAC (Stage1)
& $66.00_{\pm 6.9}$
& $\mathbf{64.67}_{\pm 9.0}$
& $\underline{73.33}_{\pm 3.1}$
& $\underline{90.00}_{\pm 4.0}$
& $86.67_{\pm 9.5}$
& $16.67_{\pm 4.6}$
& $84.00_{\pm 7.2}$
& $42.67_{\pm 10.3}$
& $76.13_{\pm 2.0}$
& $47.78_{\pm 6.3}$
& $65.50_{\pm 2.2}$ \\

\ours DexWAC (BoN)
& \rlgain{\underline{71.33}_{\pm 7.0}}
& \rldrop{59.33_{\pm 12.2}}
& \rldrop{70.67_{\pm 2.3}}
& \rldrop{86.00_{\pm 7.2}}
& \rlgain{\underline{91.33}_{\pm 4.6}}
& \rlgain{\underline{20.67}_{\pm 9.9}}
& \rlgain{\underline{86.00}_{\pm 2.0}}
& \rlgain{47.33_{\pm 1.2}}
& \rldrop{75.73_{\pm 3.5}}
& \rlgain{\underline{51.33}_{\pm 4.1}}
& \rlgain{66.58_{\pm 3.7}} \\

\ours DexWAC (PPO~\cite{schulman2017ppo})
& \rlgain{70.67_{\pm 5.8}}
& \rldrop{\underline{62.67}_{\pm 9.9}}
& \rldrop{69.33_{\pm 7.0}}
& \rlgain{\mathbf{91.33}_{\pm 4.2}}
& \rlgain{89.33_{\pm 7.0}}
& \rlgain{\mathbf{22.67}_{\pm 2.3}}
& \rlgain{\mathbf{86.67}_{\pm 3.1}}
& \rlgain{\underline{49.33}_{\pm 8.1}}
& \rlgain{\underline{76.67}_{\pm 2.3}}
& \rlgain{\mathbf{52.89}_{\pm 3.8}}
& \rlgain{\mathbf{67.75}_{\pm 1.8}} \\

\bottomrule
\end{tabular}%
}
\end{table*}

\section{Experiments}
\label{sec:experiments}

Our experiments address four questions.
(1)~Does joint world--actor--critic pretraining yield a policy competitive with state-of-the-art embodied foundation models?
(2)~Does the world model predict action consequences accurately and efficiently?
(3)~Can the adapted world model and critic further improve the pretrained
policy?
(4)~How do value supervision, visual prediction, and reward smoothing contribute to these results?

\subsection{Experimental Setup}
\label{sec:setup}

\begin{figure*}[t]
    \centering
    \includegraphics[width=\textwidth]
        {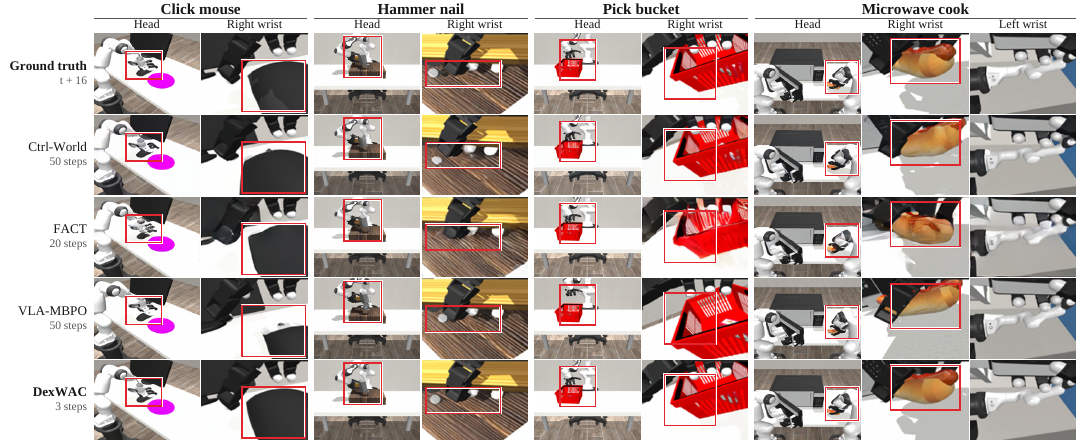}
    \caption{Qualitative comparison of action-conditioned visual prediction
    at $t+16$ on four held-out DexJoCo examples. All models receive the same
    current observations and recorded action sequence. Columns group head
    and wrist views by task; rows show ground truth and predictions from
    each model. Red boxes highlight regions for comparing gripper geometry,
    object contours, and future object positions. Ctrl-World, FACT,
    VLA-MBPO, and DexWAC use 50, 20, 50, and 3 denoising steps,
    respectively.}
    \label{fig:world-model-qualitative}
\end{figure*}

\paragraph{Benchmarks.}
Our main benchmark is DexJoCo~\cite{wang2026dexjoco}. We use eight of its
eleven tasks and exclude three that require long-horizon memory, which is
outside the scope of this study. The five single-arm tasks are
\texttt{click\_mouse}, \texttt{fold\_glasses}, \texttt{hammer\_nail},
\texttt{pick\_bucket}, and \texttt{water\_plant}; the three bimanual tasks
are \texttt{bimanual\_hanoi}, \texttt{bimanual\_microwave\_cook}, and
\texttt{bimanual\_photograph}. Together they cover diverse objects, contact
interactions, and coordination requirements. We further evaluate on the 50
tasks of RoboTwin~2.0~\cite{chen2025robotwin2} under clean and randomized settings, and on two tasks
with a physical robot (Sec.~\ref{sec:real}). 

\paragraph{Training protocol.}
On DexJoCo, we jointly pretrain a single language-conditioned
DexWAC model across all eight tasks on 800 expert demonstrations
for 10{,}000 optimizer updates, with a global batch size of 256.
We then adapt a copy of its world model and critic for
5{,}000 updates using the demonstrations and 400 policy rollouts,
including both successful and failed executions.
Policy improvement through imagination also spans all eight
tasks, using task-balanced roots to optimize a shared policy.
On RoboTwin~2.0, we jointly pretrain across all 50 tasks using
25{,}000 trajectories from the randomized setting and
2{,}500 from the clean setting. Detailed hyperparameters and implementation settings are
provided in Appendix~\ref{app:hyperparameters}.

\paragraph{Baselines and evaluation.}
We assess policy performance against the VLA model
$\pi_{0.5}$~\cite{intelligence2025pi_} and the WAMs
Motus~\cite{bi2025motus} and FACT~\cite{peng2026fact}.
For FACT, we report pretrained and Stage2 policies.
On DexJoCo, each method is evaluated on 50 episodes per task
per seed. Scenes and initial physical states are matched
across methods, while each retains its native action-sampling
procedure. On RoboTwin~2.0, we additionally include
$\pi_0$~\cite{black2024pi_0} and
Gigaworld-Policy~\cite{ni2026gigaworld}, using 100 episodes
per task in both clean and randomized settings.
To assess visual prediction separately from policy execution,
we compare with Ctrl-World~\cite{guo2026ctrlworld}, FACT,
and VLA-MBPO~\cite{zhang2026vlambpo}. All models predict the same future frames from identical
observations and actions.

\begin{table}[t]
    \centering
    \caption{Success rates (\%) across all 50 RoboTwin~2.0 tasks.
    DexWAC (BoN) uses best-of-$4$ selection on tasks where the
    pretrained actor achieves less than 85\% success and runs
    the pretrained actor directly on the remaining tasks.
    \textbf{Bold} and \underline{underline} mark the best and
    second-best results per column.}
    \label{tab:robotwin}
    \footnotesize
    \setlength{\tabcolsep}{5pt}
    \renewcommand{\arraystretch}{1.0}
    \begin{tabular}{lccc}
        \toprule
        Method & Clean & Rand. & Avg. \\
        \midrule
        $\pi_0$~\cite{black2024pi_0}           & 65.9 & 58.4 & 62.2 \\
        $\pi_{0.5}$~\cite{intelligence2025pi_} & 82.7 & 76.8 & 79.8 \\
        Gigaworld-Policy~\cite{ni2026gigaworld} & 87.0 & 85.0 & 86.0 \\
        Motus~\cite{bi2025motus}
            & \underline{88.7} & 87.0 & 87.8 \\
        FACT~\cite{peng2026fact}
            & 88.4 & 86.6 & 87.5 \\
        \midrule
        \ours DexWAC (Stage1) & 88.1 & \underline{88.2} & \underline{88.1} \\
        \ours DexWAC (BoN) & \textbf{90.1} & \textbf{90.2} & \textbf{90.2} \\
        \bottomrule
    \end{tabular}
\end{table}

\subsection{Pretrained Policy Performance}
\label{sec:pretrained}

We evaluate the jointly pretrained policy before adaptation
or policy improvement. On DexJoCo
(Table~\ref{tab:main-comparison}), DexWAC (Stage1) achieves
$65.50\pm2.2\%$ overall success across three evaluation seeds,
close to Motus ($66.83\pm3.2\%$) and above $\pi_{0.5}$
($57.92\pm5.8\%$) and both FACT variants
(49.17\% and 51.08\%).
On RoboTwin~2.0 (Table~\ref{tab:robotwin}), it achieves
the highest randomized and average success rates among
the evaluated methods, at 88.2\% and 88.1\%, respectively.
These results demonstrate competitive policy performance
from joint world--actor--critic pretraining, which also
provides dynamics and value models for subsequent improvement.

\subsection{Action-Conditioned World Modeling}
\label{sec:wm}

\begin{table}[t]
    \centering
    \caption{Action-conditioned world-model prediction quality
    at $t+16$ on 40 held-out DexJoCo trajectories (767 windows).
    \textbf{Bold} and \underline{underline} mark the best
    and second-best results.}
    \label{tab:world-model-quality}
    \setlength{\tabcolsep}{3.5pt}
    \small
    \resizebox{\columnwidth}{!}{%
    \begin{tabular}{lccccccc}
        \toprule
        & & \multicolumn{3}{c}{Head view}
        & \multicolumn{3}{c}{Wrist view} \\
        \cmidrule(lr){3-5}
        \cmidrule(lr){6-8}
        Model & Steps
        & LPIPS$\downarrow$ & PSNR$\uparrow$ & SSIM$\uparrow$
        & LPIPS$\downarrow$ & PSNR$\uparrow$ & SSIM$\uparrow$ \\
        \midrule
        Ctrl-World~\cite{guo2026ctrlworld} & 50
        & 0.0648 & 22.04 & 0.837
        & 0.1529 & 15.93 & 0.624 \\
        FACT~\cite{peng2026fact} & 20
        & \underline{0.0400} & \underline{23.51} & \underline{0.898}
        & \underline{0.1373} & \underline{16.71} & \underline{0.640} \\
        VLA-MBPO~\cite{zhang2026vlambpo} & 50
        & 0.0604 & 20.50 & 0.886
        & 0.2203 & 14.40 & 0.573 \\
        \midrule
        \ours DexWAC & 3
        & \textbf{0.0287} & \textbf{26.38} & \textbf{0.936}
        & \textbf{0.0853} & \textbf{19.01} & \textbf{0.741} \\
        \bottomrule
    \end{tabular}%
    }
\end{table}

\begin{table}[t]
\centering
\caption{End-to-end prediction latency (s) of action-conditioned
world models on DexJoCo, measured on an RTX 5090 with
batch size 1 over 24 held-out windows. \textbf{Bold}: best mean per column.}

\label{tab:world-model-latency}
\footnotesize
\setlength{\tabcolsep}{4pt}
\begin{tabular}{lrrrr}
\toprule
Model & Steps & Single-arm & Bimanual & Overall \\
\midrule
Ctrl-World~\cite{guo2026ctrlworld} & 50 & 3.875 & 5.200 & 4.372 \\
FACT~\cite{peng2026fact} & 20 & 1.800 & 1.843 & 1.816 \\
VLA-MBPO~\cite{zhang2026vlambpo} & 50 & 53.479 & 81.504 & 63.988 \\
\midrule
\ours DexWAC & 3 & \textbf{0.211} & \textbf{0.218} & \textbf{0.214} \\
\bottomrule
\end{tabular}
\end{table}

We evaluate the accuracy and efficiency of the adapted world model on DexJoCo. For a consistent comparison, DexWAC and all baselines are pretrained on the same 800 expert demonstrations, then adapted using these demonstrations and 400 policy-rollout trajectories. On held-out trajectories, all models predict observations at $t+16$ from identical observations and recorded actions. We measure prediction quality using LPIPS~\cite{zhang2018lpips}, PSNR~\cite{wang2009mse}, and SSIM~\cite{wang2004ssim} for head and wrist views, and assess efficiency through end-to-end prediction latency.

Table~\ref{tab:world-model-quality} shows that DexWAC achieves the best results across all six metrics with only three denoising steps. Improvements are particularly pronounced in wrist views, where LPIPS decreases from 0.1373 for FACT to 0.0853. The qualitative examples in Fig.~\ref{fig:world-model-qualitative} further illustrate
better preservation of gripper geometry, object contours, and future object positions. As reported in Table~\ref{tab:world-model-latency}, DexWAC predicts a chunk endpoint in 0.214\,s, approximately $8.5\times$, $20\times$, and $300\times$
faster than FACT, Ctrl-World, and VLA-MBPO, respectively.
Together, these results show that DexWAC outperforms the evaluated action-conditioned world models in both prediction quality and speed, providing an effective foundation for policy improvement.

\subsection{Model-based Policy Optimization}
\label{sec:improve}

We first test whether the adapted world model and critic can
improve action selection while keeping the pretrained actor frozen.
On RoboTwin~2.0, we use
\hyperref[par:critic-guided-bon]{critic-guided best-of-$N$}
on tasks where the pretrained actor achieves less than 85\% success.
The world model predicts the endpoint of each of four candidate
action chunks, and the critic selects the chunk with the highest
endpoint value. We use the pretrained actor directly on the remaining tasks. This raises average success across all 50 tasks from 88.1\% to 90.2\% (Table~\ref{tab:robotwin}). On DexJoCo, we adapt the world model and critic using the original
demonstrations and 50 execution trajectories per task. To reduce
test-time planning cost, we distill value estimates from imagined
transitions into a $Q$ function that ranks action chunks directly.
This raises mean success across the eight tasks from 65.50\% to
66.58\% (Table~\ref{tab:main-comparison}). We then test whether
updating the actor itself yields further gains. Starting from the
same pretrained actor, PPO on imagined rollouts raises mean success
to 67.75\%, compared with 66.83\% for Motus. In our DexJoCo
experiments, IQL~\cite{kostrikov2022iql} lowers success, and VLA-MBPO~\cite{zhang2026vlambpo} training collapses
under its original settings
(Appendix~\ref{app:additional-policy-improvement}).


\subsection{Real-Robot Experiments}
\label{sec:real}

We evaluate DexWAC on single-arm and bimanual dexterous
manipulation with two physical robots. Each task uses 300
demonstrations. Due to limited experimental resources, we evaluate DexWAC without subsequent policy improvement. For each task, we conduct 30 trials across five initial object positions to evaluate spatial generalization.

In \emph{Transfer egg}, a single arm transfers an egg
between two egg trays. In \emph{Extract cup}, the robot
uses both arms to pick up a stack of cups, extract one cup,
and place it back on the table. As shown in
Table~\ref{tab:real-robot}, DexWAC achieves 50\% and 100\%
success on the two tasks, compared with 20\% and 70\% for
$\pi_{0.5}$. These results show that DexWAC outperforms the large-scale pretrained VLA baseline on physical-robot tasks, even without subsequent policy improvement.

\begin{table}[t]
    \centering
    \caption{Success rates (\%) over 30 trials per task with randomized initial object positions. We use a Galbot G3 robot equipped with a Sharpa Dex hand on each arm. \textbf{Bold}: best mean per column.}
    \label{tab:real-robot}
    \footnotesize
    \setlength{\tabcolsep}{8pt}
    \begin{tabular}{lcc}
        \toprule
        Method & Transfer egg & Extract cup \\
        \midrule
        $\pi_{0.5}$~\cite{intelligence2025pi_} & 20 & 70 \\
        \midrule
        \ours DexWAC & \textbf{50} & \textbf{100} \\
        \bottomrule
    \end{tabular}
\end{table}

\subsection{Ablation Studies}
\label{sec:ablation}


\begin{table}[t]
\centering
\footnotesize
\caption{Pretraining-objective ablations on RoboTwin~2.0 and DexJoCo.
Success rate (\%) over 50 RoboTwin tasks and 8 DexJoCo tasks.
\textbf{Bold}: best mean per column.}
\label{tab:pretraining-objective-ablation}
\setlength{\tabcolsep}{3pt}
\begin{tabular}{@{}lcccccc@{}}
\toprule
& \multicolumn{3}{c}{RoboTwin~2.0} & \multicolumn{3}{c}{DexJoCo} \\
\cmidrule(lr){2-4}\cmidrule(lr){5-7}
Variant & Clean & Rand. & Avg. & Single & Biman. & Avg. \\
\midrule
w/o world pred.     & 82.7 & 82.8 & 82.8 & 73.20 & 52.67 & 65.50 \\
w/o reward \& cont. & --   & --   & --   & \textbf{76.40} & 49.33 & 66.25 \\
w/o critic          & 86.8 & 88.1 & 87.4 & 72.00 & 42.00 & 60.75 \\
\midrule
\ours DexWAC & \textbf{88.1} & \textbf{88.2} & \textbf{88.1}
             & \textbf{76.40} & \textbf{54.00} & \textbf{68.00} \\
\bottomrule
\end{tabular}
\end{table}

\paragraph{Actor--world--critic ablations.}
We examine how world and critic supervision contribute to
actor learning while keeping the architecture and other
training settings unchanged. On RoboTwin~2.0
(Table~\ref{tab:pretraining-objective-ablation}), removing
world prediction reduces overall success from 88.11\% to
82.8\%, while removing the critic lowers it to 87.42\%.

\begin{figure}[t]
  \centering
  \includegraphics[width=\linewidth]{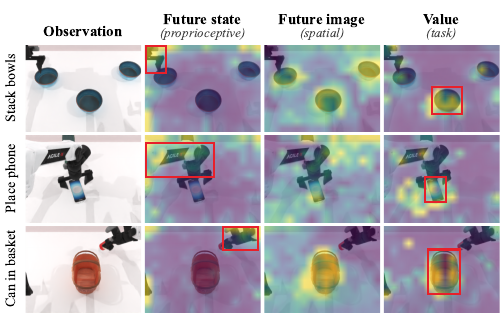}
  \caption{Attention from each prediction target to the current
head-camera tokens on RoboTwin~2.0 tasks. Future-state queries
attend to the robot arms, future-image queries spread over scene
objects, and the value token highlights task-relevant targets
(red boxes).}
  \label{fig:attention}
\end{figure}

The critic has a stronger effect on the contact-rich
DexJoCo tasks. Removing reward and continuation
prediction lowers success from 68.00\% to 66.25\%.
Removing visual prediction reduces it to
65.50\%, whereas removing the critic causes a larger drop
to 60.75\%. Dexterous manipulation often involves narrow
control bottlenecks, where small differences in contact
timing or object pose determine success. Value supervision
helps the shared representation distinguish these
task-critical states. Figure~\ref{fig:attention} visualizes
the complementary focus of world and critic supervision.
Reward smoothing is analyzed in
Appendix~\ref{app:reward-smoothing}.

\section{Limitations}
\label{sec:limitations}

Achieving sustained policy improvement in contact-rich dexterous
manipulation remains challenging. One likely reason is the difficulty of assigning credit to the actions that determine whether a contact succeeds or fails~\cite{minelli2026relative}. When a single policy is optimized across tasks, gradients from different skills can also conflict~\cite{yu2020gradient}. By keeping the actor frozen, critic-guided best-of-$N$ selection avoids conflicting actor updates during multi-task RL. It does, however, increase inference cost because the actor must sample multiple action chunks. Test-time planning with the world model further adds the cost of predicting each chunk's outcome. How to update the actor while limiting interference across tasks and retaining pretrained skills remains an open question. Finally, due to computational constraints, we have not yet evaluated DexWAC with large-scale robot pretraining. Its scalability to broader datasets, tasks, and embodiments therefore remains to be established.

\section{Conclusion}
\label{sec:conclusion}

We introduced DexWAC, an efficient framework that trains an
actor, a world model, and a critic together in a single robot
pretraining stage. The resulting policy and world model are
both competitive with strong pretrained baselines, while the
adapted world model and critic can be directly reused for
model-based policy optimization.  
Existing approaches often require the
policy and world model to be pretrained separately at large
scale, making it difficult to determine whether a failure
comes from the optimization method or from differences in
the underlying models and data. 
By providing all components
through one shared pretraining process, DexWAC reduces this
cost and enables more controlled investigation of policy
improvement. 

{\small
\bibliographystyle{ieeenat_fullname}
\bibliography{references}}
\clearpage
\appendix

\section{Hyperparameters and Implementation Settings}
\label{app:hyperparameters}

Tables~\ref{tab:supervised-settings}--\ref{tab:bon-settings} report the
settings for the DexJoCo results in Table~\ref{tab:main-comparison}.
The PPO and best-of-$N$ (BoN) variants start from the same pretrained model and use the same adaptation data, with action-value supervision enabled for BoN.

\paragraph{Shared implementation.}
We initialize the world backbone from FLUX.2-klein-base-4B and use a
width-1024 policy expert. Image autoencoder and text-encoder features are
cached. Action chunks contain $H=16$ absolute control targets; padded
action dimensions and absent camera views are masked throughout training
and inference. Optimization uses AdamW with
$\beta=(0.9,0.95)$, $\epsilon_{\mathrm{Adam}}=10^{-8}$, and global
gradient-norm clipping at 1.0. Optimizer state is initialized afresh at
each stage transition.

\subsection{Joint Pretraining and Real-Rollout Adaptation}
\label{app:supervised-settings}

Both stages use the same paired sampling formulation,
reward construction, and value-learning objective.
Real-rollout adaptation introduces execution data and
disables action imitation while retaining the predictive
and value objectives. Table~\ref{tab:supervised-settings}
summarizes the training settings.

\begin{table}[!htbp]
\centering\small
\setlength{\tabcolsep}{3pt}
\renewcommand{\arraystretch}{1.08}
\caption{Training settings for the DexJoCo comparison.
Each global batch contains 256 rows arranged as
128 adjacent anchor--endpoint pairs.}
\label{tab:supervised-settings}
\begin{tabularx}{\columnwidth}{@{}
  >{\hsize=1.2\hsize\linewidth=\hsize\raggedright\arraybackslash}X
  >{\hsize=.9\hsize\linewidth=\hsize\raggedright\arraybackslash}X
  >{\hsize=.9\hsize\linewidth=\hsize\raggedright\arraybackslash}X@{}}
\toprule
Setting & Pretraining & Adaptation \\
\midrule
Initialization & Pretrained FLUX; new expert & Stage1 checkpoint \\
Expert demonstrations & 800 & 800 \\
Policy rollouts & None & 400 \\
Optimizer updates & 10,000 & 5,000 \\
Global batch (rows) & 256 & 256 \\
Expert / rollout rows & 256 / 0 & 48 / 208 \\
GPUs $\times$ rows/GPU & $8\times32$ & $8\times32$ \\
Gradient accumulation & 1 & 1 \\
Peak learning rate & $10^{-4}$ & $10^{-5}$ \\
Warmup updates & 1,000 & 500 \\
Learning-rate decay & Cosine to $10^{-6}$ & Cosine to $10^{-6}$ \\
Weight decay & 0.01 & 0.01 \\
Flow training shift & 5 & 5 \\
\midrule
$(\lambda_A,\lambda_I,\lambda_S)$
& $(1,0.5,0.4)$ & $(0,0.5,0.4)$ \\
$(\lambda_R,\lambda_U,\lambda_V)$
& $(0.1,0.1,0.1)$ & $(0.1,0.1,0.1)$ \\
$\lambda_Q$ & 0 & 0 for PPO; 0.1 for BoN \\
Value regression & Huber, $\kappa=10$ & Huber, $\kappa=10$ \\
Reward regression & Balanced MSE & Balanced MSE \\
Reward balance $\alpha$ & 0.5 & 0.5 \\
Continuation loss & Class-balanced BCE & Class-balanced BCE \\
Reward target & Discounted interval reward & Discounted interval reward \\
TD discount & $0.99^d$ for $d$ controls & $0.99^d$ for $d$ controls \\
\bottomrule
\end{tabularx}
\end{table}

\paragraph{Reward construction.}
Both stages use the same success-preceding Gaussian
reward. For a successful trajectory ending at observation
$T$, define $D_T=\min(W_R,T)$ and
\begin{equation}
    g_t =
    \begin{cases}
        \displaystyle
        \frac{\exp[-(T-t)^2/(2\sigma^2)]}
        {\sum_{j=0}^{D_T}\exp[-j^2/(2\sigma^2)]},
        & 0\leq T-t\leq D_T, \\[6pt]
        0, & \text{otherwise}.
    \end{cases}
\end{equation}
We use $W_R=48$ and $\sigma=16$, giving a support of
at most 49 observations, including both endpoints.
Normalization is performed once per trajectory over
the available support, ensuring $\sum_t g_t=1$.
Failed trajectories have zero rewards throughout.

\paragraph{Interval rewards and value targets.}
Both stages supervise discounted interval rewards.
For an anchor $t$ and endpoint $e=\min(t+H,T)$, let
$d=e-t$. The interval reward and TD target are
\begin{align}
    r_{t:e}
    &= \sum_{j=1}^{d}
       \gamma_{\mathrm{step}}^{j-1}g_{t+j}, \\
    y_t
    &= \operatorname{sg}\!\left[
       r_{t:e}+\gamma_{\mathrm{step}}^d c_e V(C_e)
       \right],
\end{align}
where $\gamma_{\mathrm{step}}=0.99$ and $c_e$ is the
endpoint continuation label. A full action chunk has
$H=16$ and discount $0.99^{16}\approx0.8515$.

The reward head reads the endpoint context and is
conditioned on the interval length. For an ordinary
current observation at $t$, its target covers the
preceding $\min(H,t)$ controls; for a paired endpoint
at $e$, it covers exactly $(t,e]$.
Successful terminal anchors receive a zero value target,
while transitions arriving at success retain their
positive interval reward. Failed recording cutoffs
retain $c_e=1$ and may serve as bootstrap endpoints,
but are excluded from current-state supervision.

\paragraph{Paired supervision and loss balancing.}
In both stages, anchor and endpoint contexts occupy
separate rows in an ordinary batched forward pass.
The endpoint value is detached when constructing the
TD target; neither stage uses a target network or EMA.
Each global batch includes six explicitly sampled
successful terminal anchors. During adaptation,
one comes from expert data and five from rollout data.

Both stages balance reward regression identically:
positive-target and zero-target examples each contribute
one half of the loss through their respective mean
squared errors. Class counts and normalization are
computed over all valid examples in the complete
optimizer update. If either class is absent, the loss
reduces to ordinary MSE over the remaining class.

\subsection{Imagined PPO}
\label{app:ppo-settings}

\begin{table}[!htbp]
\centering\small
\setlength{\tabcolsep}{3pt}
\renewcommand{\arraystretch}{1.08}
\caption{Imagined PPO settings for DexWAC (PPO), using the expert-action
BC regularization.}
\label{tab:rl-settings}
\begin{tabularx}{\columnwidth}{@{}
  >{\raggedright\arraybackslash}X
  >{\raggedright\arraybackslash}X@{}}
\toprule
Setting & Value \\
\midrule
Initialization & Stage1 actor; Stage2 world/value model \\
Real roots per collection & 512; task-uniform, half expert / half rollout \\
Imagined horizon
& 2 chunks \\
Chunk discount & $0.99^{16}$ \\
Continuation threshold $\tau_c$ & 0.5 \\
GAE trace $\lambda$ & 0.95 \\
Policy clipping radius $\epsilon$ & 0.1 \\
Value clipping radius $\epsilon_V$ & 0.2 \\
Value regression & Clipped Huber, $\kappa=10$ \\
PPO / value / BC weights & $1 / 0.1 / 0.9$ \\
Expert BC examples per update & 128 \\
PPO global batch & 512 imagined transition rows \\
GPUs $\times$ microbatch $\times$ accumulation & $8\times32\times2$ \\
Learning rate / weight decay & $5\times10^{-7}$, constant / 0 \\
Updates per imagined buffer & At most 4 \\
Analytic KL stopping threshold & 0.02 \\
Action / world denoising steps & 3 / 3; inference shift 1 \\
Collection action sampler & Hybrid Flow-SDE; noise scale 0.5 \\
Target-value EMA / PCGrad & Neither \\
Reported checkpoint & 500 updates \\
\bottomrule
\end{tabularx}
\end{table}

Both FLUX backbones remain frozen. The actor expert receives PPO and
expert-action flow-matching BC gradients; the world expert receives
value gradients only. BC examples are drawn from legal expert roots
in the collected buffer and use recorded actions, without a teacher
policy or rollout-action imitation.

The hybrid sampler uses one selected stochastic denoising transition
per sampled chain. Its behavior log-probability is stored during
collection; optimization recomputes only the current-policy
log-probability on that same saved transition. The likelihood sums
over valid action coordinates. The stopping statistic is the
full-buffer mean analytic Gaussian transition KL
$D_{\mathrm{KL}}(p_{\mathrm{old}}\Vert p_\theta)$, checked after each
optimizer update. Crossing 0.02 ends buffer reuse without rolling back
the completed update. Predicted interval rewards are consumed directly,
without an additional within-chunk discount.

\subsection{Best-of-$N$ Critic Training and Deployment}
\label{app:bon-settings}

\begin{table}[!htbp]
\centering\small
\setlength{\tabcolsep}{3pt}
\renewcommand{\arraystretch}{1.08}
\caption{Critic training and proposal selection for DexWAC (BoN).}
\label{tab:bon-settings}
\begin{tabularx}{\columnwidth}{@{}
  >{\raggedright\arraybackslash}X
  >{\raggedright\arraybackslash}X@{}}
\toprule
Setting & Value \\
\midrule
Frozen components & Entire Stage1 actor and Stage2 world model \\
Critic initialization & Adapted Stage2 expert, including $V$ and $Q$ heads \\
Critic size / tokens & 644M parameters / $V+16$ action tokens $+Q$ \\
Real roots per update & 512; task-uniform, half expert / half rollout \\
Imagined horizon / candidates & 1 chunks / 8 proposals \\
Deployment candidates $N$ & 8 \\
Candidate selection & Maximum online $Q$ \\
Chunk discount & $0.99^{16}$ \\
Continuation threshold & 0.5 \\
$Q$ bootstrap & Detached online successor $V$ \\
Multi-step $V$ tail & Detached EMA successor $V$ \\
Critic EMA rate & 0.02 per optimizer update \\
Critic loss & Scalar MSE; $Q$ and $V$ weights 0.1 each \\
Learning rate / weight decay & $5\times10^{-7}$, constant / 0 \\
Action / world denoising steps & 3 / 3; inference shift 1 \\
Reported checkpoint & 500 updates \\
\bottomrule
\end{tabularx}
\end{table}

The shared critic expert and both $V$ and $Q$ heads are
initialized from Stage2 and jointly optimized on imagined
transitions from the first update. Each imagined rollout
uses two action chunks, with actions selected from eight
frozen-actor proposals. Real roots exclude terminal states.

The EMA critic is initialized from the adapted critic
and updated as
$\bar\phi\leftarrow0.98\bar\phi+0.02\phi$.
It supplies the tail value for multi-chunk returns,
while the $Q$ target uses the online successor value.


\section{Action-Generation Latency}
\label{app:latency}

\begin{table}[t]
\centering
\small
\caption{End-to-end action-generation latency on RTX 5090
GPUs with batch size 1. Each method is measured over
80 calls: 10 per task across eight tasks.
Timing includes preprocessing and image encoding,
but excludes model loading and simulator rendering.
Methods retain their native evaluation configurations.
BoN includes all 32 candidate generations and critic ranking.
Bold indicates the lowest latency in each column.}
\label{tab:latency}
\setlength{\tabcolsep}{3pt}
\renewcommand{\arraystretch}{1.1}
\begin{tabular}{@{}lcccrr@{}}
\toprule
Method & GPUs & Steps & Chunk
& \shortstack{Mean\\(ms)} & \shortstack{P95\\(ms)} \\
\midrule
$\pi_{0.5}$     & 1 & 10 & 30 & 352.4  & 357.5 \\
Motus          & 2 & 3  & 16 & 626.7  & 634.1 \\
FACT (Stage1)  & 1 & 20 & 48 & 2665.4 & 2727.0 \\
FACT (Stage2)  & 1 & 20 & 48 & 2675.8 & 2730.1 \\
\midrule
DexWAC (Stage1) & 1 & 3 & 16
& \textbf{115.9} & \textbf{119.8} \\
DexWAC (PPO)   & 1 & 3 & 16 & 118.0  & 123.3 \\
DexWAC (BoN)   & 1 & 3 & 16 & 3116.0 & 3156.7 \\
\bottomrule
\end{tabular}
\end{table}

DexWAC requires 115.9\,ms per action chunk after pretraining
and 118.0\,ms after PPO optimization.
BoN increases latency to 3116.0\,ms, approximately
$26.9\times$ that of direct Stage1 sampling, reflecting
the cost of generating and ranking 32 candidates.
These measurements characterize deployment latency under
the evaluated configurations rather than matched-compute
comparisons.

\section{Policy Improvement}
\label{app:policy-improvement}

We use two existing policy-improvement approaches:
flow-based PPO following $\pi_{\mathrm{RL}}$
\cite{chen2025pirl}, and critic learning from imagined
transitions following MAC~\cite{park2026mac}.
The actor is initialized from pretraining and the critic
from real-rollout adaptation. Below, $\psi$ and $\omega$
denote the actor- and critic-expert parameters, respectively.
Their backbones remain frozen.

\paragraph{Imagined transitions.}
For each imagined chunk transition, we use
\begin{equation}
    R_j=\widehat R(\widehat C_{j+1},H),
    \qquad
    b_j=\gamma_{\mathrm{step}}^H\widehat c_j.
    \label{eq:app-chunk-discount}
\end{equation}
Here, $R_j$ already includes discounting within the chunk,
while $b_j$ discounts subsequent returns and removes
bootstrapping upon termination. All objectives use valid
transitions only, excluding padded steps.

\subsection{PPO for Flow-Based Policies}

\paragraph{Flow-policy update.}
Following $\pi_{\mathrm{RL}}$~\cite{chen2025pirl}, we use
a hybrid denoising sampler with one selected stochastic
transition per action chunk. Its Gaussian conditional
density provides a tractable likelihood for PPO:
\begin{equation}
    \rho_j(\psi)=
    \frac{
      p_\psi(x_{\tau'}\mid x_\tau,\tau,\widehat C_j)
    }{
      p_{\mathrm{old}}(x_{\tau'}\mid x_\tau,\tau,\widehat C_j)
    }.
    \label{eq:app-ratio}
\end{equation}
Here, $x_\tau$ and $x_{\tau'}$ are the input and output
of the sampled denoising transition. The ratio is defined
for this transition rather than the final action chunk.
We retain the sampled transition and behavior statistics
throughout buffer reuse, and evaluate likelihoods over
valid action coordinates.

We optimize the clipped PPO objective with the expert-action
BC regularization in Eq.~\eqref{eq:ppo-anchor}. Gradients are computed
through the selected transition likelihood and the BC loss;
the sampled denoising chain and imagined dynamics are not
differentiated through. We limit updates per buffer and use
the mean conditional KL between behavior and updated
transitions for early stopping.

\paragraph{Advantages and critic updates.}
We compute chunk-level advantages using
GAE~\cite{schulman2016gae}:
\begin{equation}
\begin{aligned}
    \delta_j
        &=R_j+b_jV_{j+1}^{\mathrm{ref}}
          -V_j^{\mathrm{ref}},\\
    \widehat A_j
        &=\delta_j+\lambda b_j\widehat A_{j+1},\\
    y_j^V
        &=\operatorname{sg}
          [V_j^{\mathrm{ref}}+\widehat A_j],
\end{aligned}
\label{eq:app-gae}
\end{equation}
where $V_j^{\mathrm{ref}}$ denotes a detached value
prediction and $\widehat A_K=0$ at the rollout boundary.
Nonterminal endpoints retain their value bootstrap.
The world-side critic is trained by Huber regression
to $y_j^V$, separately from the actor expert.
Imagined states, rewards, advantages, and value targets
are detached from actor optimization.

\subsection{Critic-Guided Best-of-$N$}

\paragraph{Critic initialization.}
During real-rollout adaptation, the critic jointly learns
$V_\omega(C)$ and $Q_\omega(C,\mathbf a)$ from recorded
transitions. Both use the detached target in
Eq.~\eqref{eq:valuetarget}. The value route reads only
the observed context, while the $Q$ route additionally
reads the clean action chunk.

\paragraph{Learning from imagination.}
Following MAC~\cite{park2026mac}, we train state values
on multi-chunk returns and action values on one-chunk
bootstrap targets:
\begin{equation}
\begin{aligned}
    G_K &= V_{\bar\omega}(\widehat C_K),\\
    G_j &= R_j+b_jG_{j+1},\\
    y_j^V &= \operatorname{sg}[G_j],\\
    y_j^Q &= \operatorname{sg}
        [R_j+b_jV_\omega(\widehat C_{j+1})],
\end{aligned}
\label{eq:app-mac-targets}
\end{equation}
where $\bar\omega$ denotes an exponential-moving-average
copy of the critic. The online value function supplies
the $Q$ bootstrap, while the EMA value is used at the
tail of the multi-chunk return.

The shared critic expert and both prediction heads
minimize
\begin{equation}
\begin{aligned}
    \mathcal L_{\mathrm{critic}}
    ={}&\lambda_V\mathbb E_j
       [(V_\omega(\widehat C_j)-y_j^V)^2]\\
    &+\lambda_Q\mathbb E_j
       [(Q_\omega(\widehat C_j,\mathbf a_j)-y_j^Q)^2].
\end{aligned}
\label{eq:app-joint-critic}
\end{equation}
The EMA parameters are updated as
$\bar\omega\leftarrow
(1-\tau_{\mathrm{EMA}})\bar\omega+\tau_{\mathrm{EMA}}\omega$.
The pretrained actor and adapted world model remain
frozen throughout critic training.

\paragraph{Action selection.}
At deployment, we sample $N=32$ action chunks from the
frozen actor and select the highest-scoring proposal
using Eq.~\eqref{eq:bon}. Context features are reused
across candidates. The world model supplies imagined
transitions during critic training but is not evaluated
during test-time ranking.

\subsection{Additional Experiments}
\label{app:additional-policy-improvement}

We additionally evaluate direct offline policy improvement
with IQL~\cite{kostrikov2022iql} and the original
VLA-MBPO optimization settings~\cite{zhang2026vlambpo}.
Table~\ref{tab:additional-policy-improvement} reports
the results on the eight DexJoCo tasks.

\begin{table*}[t]
\centering
\caption{
Additional policy-improvement results on DexJoCo.
Entries report success rates (\%).
PPO results correspond to evaluation seed~0 after
500 updates. Dashes indicate that the original
VLA-MBPO settings led to policy collapse and did not
yield a success rate.
}
\label{tab:additional-policy-improvement}
\setlength{\tabcolsep}{3pt}
\renewcommand{\arraystretch}{1.12}
\resizebox{\textwidth}{!}{%
\begin{tabular}{@{}l*{11}{c}@{}}
\toprule
& \multicolumn{5}{c}{Single-arm tasks}
& \multicolumn{3}{c}{Bimanual tasks}
& \multicolumn{3}{c}{Average} \\
\cmidrule(lr){2-6}
\cmidrule(lr){7-9}
\cmidrule(lr){10-12}
Method
& \shortstack{Click\\mouse}
& \shortstack{Fold\\glasses}
& \shortstack{Hammer\\nail}
& \shortstack{Pick\\bucket}
& \shortstack{Water\\plant}
& Hanoi
& \shortstack{Microwave\\cook}
& Photograph
& Single-arm & Bimanual & Overall \\
\midrule
DexWAC (Stage1)
& 62.00 & 74.00 & 70.00 & 86.00 & 90.00
& 22.00 & 86.00 & 54.00
& 76.40 & 54.00 & 68.00 \\

IQL~\cite{kostrikov2022iql}
& 56.00 & 40.00 & 46.00 & 100.00 & 60.00
& 26.00 & 80.00 & 30.00
& 60.40 & 45.33 & 54.75 \\

VLA-MBPO~\cite{zhang2026vlambpo}
& -- & -- & -- & -- & --
& -- & -- & --
& -- & -- & -- \\
\midrule
\ours DexWAC (PPO~\cite{schulman2017ppo})
& 74.00 & 74.00 & 62.00 & 90.00 & 90.00
& 24.00 & 86.00 & 58.00
& 78.00 & 56.00 & 69.75 \\
\bottomrule
\end{tabular}%
}
\end{table*}

\paragraph{Direct offline policy improvement.}
Direct application of IQL yields 54.75\% overall success,
below the pretrained policy's 68.00\%.
Although some tasks improve, these gains are outweighed
by regressions on others. We attribute this degradation
in part to unstable Q-value estimates under limited
data coverage. In contact-rich manipulation, small
action differences can change the outcome substantially,
making reliable value estimation difficult.
Errors in the learned values can consequently steer
policy updates away from effective pretrained behavior.

\paragraph{Optimization with the original VLA-MBPO settings.}
Applying the original VLA-MBPO optimization settings
in our setup leads to policy collapse. We attribute
this behavior to accumulated world-model errors during
imagined rollouts. Inaccurate contact predictions produce
unreliable future states and reward targets, which
optimization can exploit rather than improve actual
execution. These observations highlight the difficulty
of directly transferring existing optimization settings
to contact-rich dexterous manipulation with limited
execution data.

\section{Reward-Target Smoothing}
\label{app:reward-smoothing}

We examine whether smoothing the reward targets helps the
model recognize successful terminal states. We compare a
checkpoint trained with binary targets and binary
cross-entropy against one trained with Gaussian-smoothed
targets and class-balanced mean squared error. Both
checkpoints are evaluated on the same 400 trajectories
collected by Motus with evaluation seed~1. These trajectories
contain 266 successful terminal observations across eight
DexJoCo tasks.

A successful terminal observation is counted as detected
when its predicted reward exceeds half of the corresponding
target: $0.5$ for the binary target and half of the
normalized Gaussian terminal value for the smoothed target.
No threshold calibration is performed.

\begin{figure}[t]
    \centering
    \includegraphics[width=\linewidth]
        {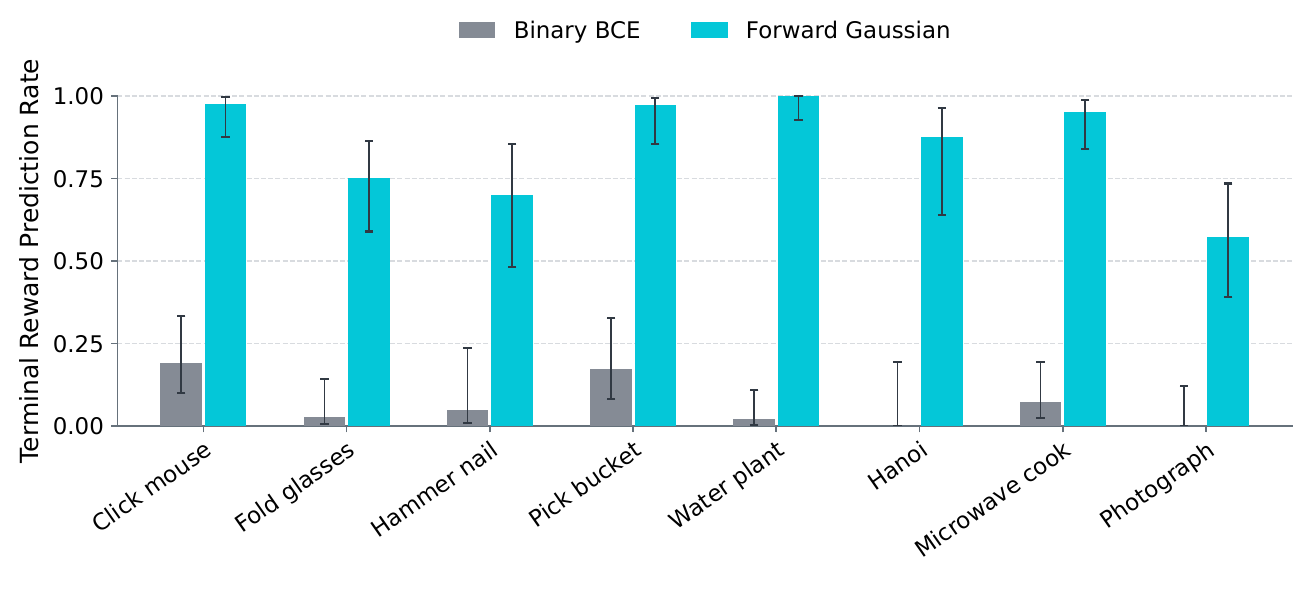}
    \caption{Terminal reward detection across eight DexJoCo
    tasks. Detection rates are computed over successful
    episodes using a half-target threshold. Error bars show
    95\% Wilson confidence intervals.}
    \label{fig:reward-prediction}
\end{figure}

As shown in Fig.~\ref{fig:reward-prediction}, the checkpoint
trained with Gaussian-smoothed targets achieves higher
terminal detection rates on all eight tasks. This diagnostic
measures recall on successful terminal states and does not
evaluate false positives or downstream policy performance.
Because the checkpoints also differ in initialization and
other training settings, the comparison should not be
interpreted as isolating the effect of reward smoothing
alone.

\begin{table*}[t]
\centering
\caption{
Per-seed success rates (\%) on the eight-task DexJoCo subset
under deterministic evaluation, with 50 episodes per task
per seed. PPO uses an expert-action BC regularization, and BoN uses
a MAC-trained critic for best-of-32 selection; both are
evaluated after 500 updates.
Summary rows report means with sample standard deviations
across three seeds.
Green and red indicate gains and drops relative to Stage1
within each seed or in the mean; blue indicates Stage1
or unchanged performance.
Bold and underline in the summary rows indicate the best
and second-best means among all methods in
Table~\ref{tab:main-comparison}.
}
\label{tab:dexwac-per-seed}
\setlength{\tabcolsep}{3pt}
\renewcommand{\arraystretch}{1.15}
\resizebox{\textwidth}{!}{%
\begin{tabular}{@{}ll*{11}{c}@{}}
\toprule
& & \multicolumn{5}{c}{Single-arm tasks}
& \multicolumn{3}{c}{Bimanual tasks}
& \multicolumn{3}{c}{Average} \\
\cmidrule(lr){3-7}
\cmidrule(lr){8-10}
\cmidrule(lr){11-13}
Seed & Method
& \shortstack{Click\\mouse}
& \shortstack{Fold\\glasses}
& \shortstack{Hammer\\nail}
& \shortstack{Pick\\bucket}
& \shortstack{Water\\plant}
& Hanoi
& \shortstack{Microwave\\cook}
& Photograph
& Single-arm & Bimanual & Overall \\
\midrule

\rowcolor{OursBG}
\cellcolor{white}
& \cellcolor{white}DexWAC (Stage1)
& $62.00$ & $74.00$ & $70.00$ & $86.00$ & $90.00$
& $22.00$ & $86.00$ & $54.00$
& $76.40$ & $54.00$ & $68.00$ \\

\rowcolor{OursBG}
\cellcolor{white}
& \cellcolor{white}DexWAC (PPO~\cite{schulman2017ppo})
& \rlgain{74.00}
& $74.00$
& \rldrop{62.00}
& \rlgain{90.00}
& $90.00$
& \rlgain{24.00}
& $86.00$
& \rlgain{58.00}
& \rlgain{78.00}
& \rlgain{56.00}
& \rlgain{69.75} \\

\rowcolor{OursBG}
\cellcolor{white}\multirow{-3}{*}{0}
& \cellcolor{white}DexWAC (BoN~\cite{BoN})
& \rlgain{72.00}
& \rldrop{70.00}
& \rldrop{68.00}
& \rlgain{94.00}
& \rlgain{94.00}
& \rlgain{32.00}
& \rlgain{88.00}
& \rldrop{48.00}
& \rlgain{79.60}
& \rlgain{56.00}
& \rlgain{70.75} \\
\midrule

\rowcolor{OursBG}
\cellcolor{white}
& \cellcolor{white}DexWAC (Stage1)
& $62.00$ & $64.00$ & $76.00$ & $94.00$ & $94.00$
& $14.00$ & $76.00$ & $34.00$
& $78.00$ & $41.33$ & $64.25$ \\

\rowcolor{OursBG}
\cellcolor{white}
& \cellcolor{white}DexWAC (PPO~\cite{schulman2017ppo})
& \rlgain{64.00}
& \rldrop{58.00}
& $76.00$
& \rlgain{96.00}
& \rlgain{96.00}
& \rlgain{20.00}
& \rlgain{84.00}
& \rlgain{42.00}
& $78.00$
& \rlgain{48.67}
& \rlgain{67.00} \\

\rowcolor{OursBG}
\cellcolor{white}\multirow{-3}{*}{1}
& \cellcolor{white}DexWAC (BoN~\cite{BoN})
& \rlgain{64.00}
& \rldrop{62.00}
& \rldrop{72.00}
& \rldrop{80.00}
& \rldrop{86.00}
& \rlgain{16.00}
& \rlgain{84.00}
& \rlgain{46.00}
& \rldrop{72.80}
& \rlgain{48.67}
& \rldrop{63.75} \\
\midrule

\rowcolor{OursBG}
\cellcolor{white}
& \cellcolor{white}DexWAC (Stage1)
& $74.00$ & $56.00$ & $74.00$ & $90.00$ & $76.00$
& $14.00$ & $90.00$ & $40.00$
& $74.00$ & $48.00$ & $64.25$ \\

\rowcolor{OursBG}
\cellcolor{white}
& \cellcolor{white}DexWAC (PPO~\cite{schulman2017ppo})
& $74.00$
& $56.00$
& \rldrop{70.00}
& \rldrop{88.00}
& \rlgain{82.00}
& \rlgain{24.00}
& $90.00$
& \rlgain{48.00}
& $74.00$
& \rlgain{54.00}
& \rlgain{66.50} \\

\rowcolor{OursBG}
\cellcolor{white}\multirow{-3}{*}{2}
& \cellcolor{white}DexWAC (BoN~\cite{BoN})
& \rlgain{78.00}
& \rldrop{46.00}
& \rldrop{72.00}
& \rldrop{84.00}
& \rlgain{94.00}
& $14.00$
& \rldrop{86.00}
& \rlgain{48.00}
& \rlgain{74.80}
& \rlgain{49.33}
& \rlgain{65.25} \\
\midrule

\rowcolor{OursBG}
\cellcolor{white}
& \cellcolor{white}DexWAC (Stage1)
& $66.00_{\pm 6.9}$
& $\mathbf{64.67}_{\pm 9.0}$
& $\underline{73.33}_{\pm 3.1}$
& $\underline{90.00}_{\pm 4.0}$
& $86.67_{\pm 9.5}$
& $16.67_{\pm 4.6}$
& $84.00_{\pm 7.2}$
& $42.67_{\pm 10.3}$
& $76.13_{\pm 2.0}$
& $47.78_{\pm 6.3}$
& $65.50_{\pm 2.2}$ \\

\rowcolor{OursBG}
\cellcolor{white}
& \cellcolor{white}DexWAC (PPO~\cite{schulman2017ppo})
& \rlgain{70.67_{\pm 5.8}}
& \rldrop{\underline{62.67}_{\pm 9.9}}
& \rldrop{69.33_{\pm 7.0}}
& \rlgain{\mathbf{91.33}_{\pm 4.2}}
& \rlgain{89.33_{\pm 7.0}}
& \rlgain{\mathbf{22.67}_{\pm 2.3}}
& \rlgain{\mathbf{86.67}_{\pm 3.1}}
& \rlgain{\underline{49.33}_{\pm 8.1}}
& \rlgain{\underline{76.67}_{\pm 2.3}}
& \rlgain{\mathbf{52.89}_{\pm 3.8}}
& \rlgain{\mathbf{67.75}_{\pm 1.8}} \\

\rowcolor{OursBG}
\cellcolor{white}\multirow{-3}{*}{\shortstack{Mean\\$\pm$ SD}}
& \cellcolor{white}DexWAC (BoN~\cite{BoN})
& \rlgain{\underline{71.33}_{\pm 7.0}}
& \rldrop{59.33_{\pm 12.2}}
& \rldrop{70.67_{\pm 2.3}}
& \rldrop{86.00_{\pm 7.2}}
& \rlgain{\underline{91.33}_{\pm 4.6}}
& \rlgain{\underline{20.67}_{\pm 9.9}}
& \rlgain{\underline{86.00}_{\pm 2.0}}
& \rlgain{47.33_{\pm 1.2}}
& \rldrop{75.73_{\pm 3.5}}
& \rlgain{\underline{51.33}_{\pm 4.1}}
& \rlgain{66.58_{\pm 3.7}} \\
\bottomrule
\end{tabular}%
}
\end{table*}

\section{Discussion}
\label{app:discussion}

\paragraph{Multi-seed policy improvement.}
Table~\ref{tab:dexwac-per-seed} reports the complete
DexJoCo results across three evaluation seeds.
PPO provides the clearest evidence of multi-task policy
improvement: it improves overall success on every seed
and raises the mean from 65.50\% to 67.75\% after
500 updates. Most of the gain comes from bimanual tasks,
whose average success increases from 47.78\% to 52.89\%.
However, the improvement is not uniform across tasks.
Several skills improve substantially, while others exhibit
modest regressions. Optimizing the shared policy therefore
improves its average performance without guaranteeing
retention of every pretrained skill.

\paragraph{Over-optimization in imagination.}
Continued optimization does not necessarily produce further
gains. With only 50 execution trajectories per task, the
adapted world model cannot cover every contact outcome that
arises during imagined exploration. As training proceeds,
the policy or critic may increasingly exploit prediction
and reward errors, favoring imagined outcomes that receive
high predicted returns but are not physically realizable.
This form of model exploitation is especially pronounced
for BoN, where maximizing over multiple proposals amplifies critic ranking errors. 

\paragraph{Sensitivity of contact-rich manipulation.}
These results expose a central difficulty of model-based
policy improvement for contact-rich manipulation.
Small differences in pose, timing, or contact force can
turn a successful grasp or insertion into a slip, collision,
or missed contact. Such differences may appear minor in
predicted observations but correspond to different task
outcomes. Errors in imagined dynamics can then propagate
into reward and value targets and become amplified through
policy optimization or best-of-$N$ selection. More reliable
improvement will require better uncertainty estimates,
conservative critic updates, validation on real executions,
and mechanisms for stopping optimization before the policy
or critic begins to exploit model errors.

\end{document}